\documentclass[sigconf]{acmart}
\AtBeginDocument{%
  }
\copyrightyear{2025}
\acmYear{2025}
\acmConference[SIGIR '25]{The ACM Special Interest Group on Information Retrieval}{July 13--18,
  2025}{Padua, Italy}
\acmSubmissionID{14881.311}

\usepackage{amsmath}
\usepackage{amsfonts}
\usepackage{graphicx}
\usepackage{multirow}
\usepackage{tabularx}
\usepackage{subcaption}

\begin{document}

\title{EIoU-EMC: A Novel Loss for Domain-specific Nested Entity Recognition }

\author{Jian Zhang}
\email{12221038@zju.edu.cn}
\orcid{0000-0002-6342-0243}
\affiliation{%
  \institution{College of Computer Science and Technology, Zhejiang University}
  \city{Hangzhou}
  \state{Zhejiang}
  \country{China}
}
\author{Tianqing Zhang}
\email{tianqing21@intl.zju.edu.cn}
\orcid{0009-0003-5500-1828}
\affiliation{%
  \institution{College of Computer Science and Technology, Zhejiang University}
  \city{Hangzhou}
  \state{Zhejiang}
  \country{China}
}
\author{Qi Li}
\email{qili177@zju.edu.cn}
\orcid{0009-0006-4729-0289}
\affiliation{%
  \institution{College of Computer Science and Technology, Zhejiang University}
  \city{Hangzhou}
  \state{Zhejiang}
  \country{China}
}

\author{Hongwei Wang}
\email{hongweiwang@zju.edu.cn}
\orcid{0000-0001-6118-6505}
\authornote{Corresponding Author.}
\affiliation{%
  \institution{ZJU-UIUC Institute, Zhejiang University}
  \city{Haining}
  \state{Zhejiang}
  \country{China}
}

\renewcommand{\shortauthors}{Zhang et al.}

\begin{abstract}
 In recent years, research has mainly focused on the general NER task. There still have some challenges with nested NER task in the specific domains. Specifically, the scenarios of low resource and class imbalance  impede the wide application for biomedical and industrial domains.
In this study, we design a novel loss EIoU-EMC, by enhancing the implement of Intersection over Union loss and Multi-class loss. 
Our proposed method specially leverages the information of  entity boundary and entity classification, thereby enhancing the model's capacity to learn from a limited number of data samples.
To validate the performance of this innovative method in enhancing NER task, we conducted experiments on three distinct biomedical NER datasets and one dataset constructed by ourselves from industrial complex equipment maintenance documents. Comparing to strong baselines, our method   demonstrates the competitive performance   across all datasets. During the experimental analysis, our proposed method exhibits significant advancements in entity boundary recognition and entity classification. 
Our code and data are available at https://github.com/luminous11/EIoU-EMC/
\end{abstract}



\maketitle

\section{Introduction}

As a fundamental task in Natural Language Processing (NLP), Named Entity Recognition \cite{tjongkimsangIntroductionCoNLL2003Shared2003} (NER) aims to extract entities from unstructured text and documents, contributing to the construction of knowledge graphs and enhancing the performance of downstream application in information retrieval applications.
Over recent years, numerous deep learning models have been employed for nested NER \cite{finkelNestedNamedEntity2009}. In contrast to flat NER, where entities are independent \cite{xuNamedEntityRecognition2019},  
the nested entities exhibit hierarchical structure. Considering both entity boundaries and categories, identifying overlapping entities is a more challenging task. Previous works on nested NER mainly fall into categories such as layer-based \cite{feiDispatchedAttentionMultitask2020}, 
hypergraph-based \cite{wangNeuralSegmentalHypergraphs2018} 
and region-based \cite{yangTsERLTwostageEnhancement2022} 
methods.
\begin{figure}
    \centering
    \includegraphics[width=0.9\columnwidth]{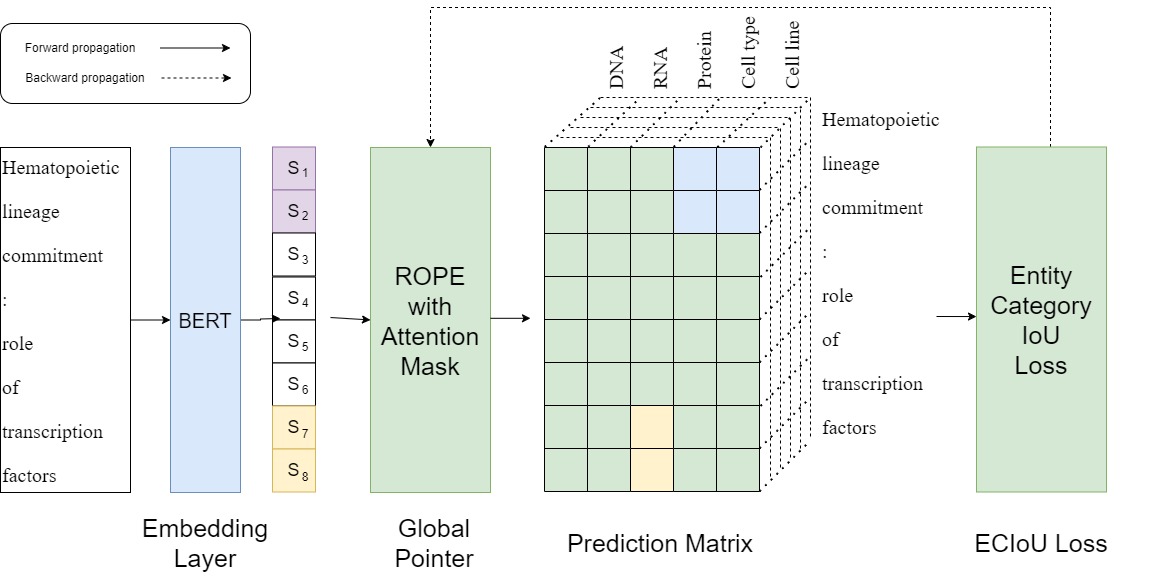}
    \caption{\textbf{A sample of entity bounding box.} 
    We devised a process for extracting entity boundaries and categories from the prediction matrix. 
    This procedure can be regarded as akin to locating bounding boxes in a photograph, wherein an EIoU-EMC loss function is established to compute the distance between the prediction and the golden true matrix.}
    \label{fig:bciou}
\end{figure}
However, these methods struggle with low-resource and long-tail imbalanced problems, which are common challenges when applying nested NER to specific domains. For the low resource problem  \cite{xuMultiTaskInstructionChain2023}, the primary reason is the lack of sufficient domain-specific corpus or datasets for the training process. Thus, the model cannot accurately capture the features of entity types in the specific domains.
Addressing the challenge of long-tail imbalanced classification, the distribution of different entity types varies significantly in specific domains such as medical or industrial fields. 
This imbalance complicates the learning process for the NER model.
Inspired by the idea of metrics of bounding box regression in 2D computer vision tasks, we propose Entity Intersection over Union (EIoU) loss to address the challenges of low resources by enhancing the correlation between entity spans across different classes. By integrating Entity Multi-class Cross-entropy (EMC) loss, we propose the EIoU-EMC loss function, which improves the performance on imbalanced classification problems, significantly boosting the metrics for minority classes.

The contributions can be summarized as follows:

\textbf{Introduction of EIoU-EMC loss.}
We introduce a novel loss function, EIoU-EMC, alleviating the issues of low resources and class imbalance within nested NER task for information retrieval applications.

\textbf{Enhanced performance across NER task.}
Leveraging several strong baselines, we show the high performance of our proposed method in three public NER datasets and a newly constructed application dataset. 
Specially, this dataset with characteristics of low resources and class imbalance is firstly introduced in this work.

\textbf{Comprehensive analysis on our method.}
Experimental analysis confirmed the excellent performance of the proposed loss function in both the boundary recognition subtask and the entity classification subtask in the scenarios of low resource and class imbalance.

\section{Related Work}

\subsection{Low Resource NER}
The problem of low resource training in deep learning is common in various domains. In the case of NER, \citet{fuSelfTrainingDoubleSelectors2023} focus on the pseudo sample selection and a self-training framework with double selectors to alleviating insufficient data. \citet{nguyenAUCMaximizationLowResource2023} proposed an AUC maximizing strategy to optimize the NER model on low resource scenario. The prompt-based text entailment method \cite{tanPromptEnhancedGenerative2022} is apply for low resource NER. In addition, there are also some works \cite{xuCanNLIProvide2023, caoLowResourceNameTagging2019,leeEnhancingLowresourceFinegrained2023} are focus on low resource NER.

\subsection{Imbalanced Classification}
In the context of imbalanced classification for NER, \citet{liDiceLossDataimbalanced2020} proposed to a dice loss for data-imbalanced NLP task to alleviate the domainating influence from easy-negative examples in training. \citet{shenLocateLabelTwostage2021b} utilized boundary information of entities and partially mathced spans during training by a two-stage entity identifier. \citet{jiangCandidateRegionAware2021} designed a multi-tasks model to binary sequence labeling and candidate region classification. 

\section{Proposed Method}

We propose a novel loss function to solve the problem in the domain-specific nested NER task, for addressing the issues of low resource and imbalanced entity quantity. The framework of our proposed method is shown in Fig. \ref{fig:bciou}.

\subsection{Task Definition} \label{sec:def}

NER task aims to extract entities of pre-defined types in a sentence \cite{finkelNestedNamedEntity2009}. For a sentence  $X=\{x_1, x_2, \ldots, x_i, \ldots, x_j, \ldots, x_k\}$ with $k$ tokens,  the set of  all  spans is denoted as $S = \{s_{i:j}\}$, where the $i$ and $j$ are the start token and end token of the span, respectively. 
The goal of the  NER task  is to identify all entities  in  $S$ and classify these entities  into  the pre-defined types. Formally, for a given span $s_{i:j}$, the NER model is trained to predict their entity type $y_{i:j} \in E$.

\subsection{Span Representation} \label{sec:rep}

Typically, the span representation  $s_{i:j}=h_{i}^{\top} t_{j} $ mainly considers the start token $x_i$ and end token $x_j$, and these components of  $h_{i} $ and $t_{j} $ are  defined as follows:

\begin{equation}
\renewcommand{\arraystretch}{0.8}
\begin{aligned}
      h_{i} = g_s(f(x_i)),\\
      t_{j} = g_e(f(x_j)). \\
\end{aligned}
\end{equation}
The function $f$ represents the pre-trained BERT-based language model to obtain the token representation. The function $ g_{s}$ and $ g_{e}$ serve as  feedback layers for the start token representation and the end token representation, respectively. 
 
In advance, we refer to the previous work  \cite{suGlobalPointerNovel2022} to better utilize boundary information. Therefore, the span representation is formatted as:

\begin{equation}
\renewcommand{\arraystretch}{0.8}
\begin{aligned}
    s_{i:j}  = (\mathrm{R}_i h_{i})^\top(\mathrm{R}_j t_{j})   = h_i^\top \mathrm{R}_i^\top \mathrm{R}_j t_j= h_i^\top \mathrm{R}_{j-i} t_j, 
\end{aligned}
\end{equation}
where $\mathrm{R}_i=\text{RoPE}(h_i)$ and $\mathrm{R}_j=\text{RoPE}(t_j)$.

\subsection{Loss Function} \label{sec:function}

Compared to the general domain, the NER task in medical or industrial fields presents specific challenges. First, collecting a training dataset with sufficient samples for each entity type is difficult. Enhancing the model's learning ability is essential to enable it to learn features from limited data. Additionally, imbalanced data across entity types complicates the learning process for the NER model. It is crucial to optimize the model to handle imbalanced samples, ensuring it can accurately recognize entity types with fewer samples.

We introduce two different loss functions to address two distinct issues. First, Entity Intersection over Union loss, leverages entity category and entity length to enhance the learning process for entity features. Second, Entity Multi-Class Imbalance loss, the second function is derived from imbalanced classification task and address the issue of multi label class imbalance in the scenarios. 

\paragraph{Entity IoU Loss.}

Inspired by Intersection over Union (IoU) loss \cite{yuUnitBoxAdvancedObject2016, rezatofighiGeneralizedIntersectionUnion2019, zhengEnhancingGeometricFactors2022} for object detection, we proposed an Entity IoU Loss (EIoU) for nested entity recognition task. 
The process of identifying entities in NER is analogous to locating objects in traditional object detection. Thus, we treat the entity area as  the bounding box rectangle, as shown in Fig. \ref{fig:bciou}. 
In this context, the width of the bounding box rectangle corresponds to the length of the entity in the NER task, and the height of the bounding box rectangle is set to 1 in the NER task. The detailed definition of the EIoU loss is provided below.

\begin{equation}
\renewcommand{\arraystretch}{0.8}
\begin{aligned}
 V = &\frac{4}{\pi} ( \arctan \frac{b^{gt}}{c^{gt}} - \arctan \frac{b}{c})^2, \\
    \mathcal{L}_{EIoU} =& 1 + \ln{\frac{A \cap A^{gt}}{A \cup A^{gt}}} + \frac{\rho^2(C, C^{gt})}{b^2+c^2} + \frac{V^2}{(1-IoU) + V}, \\
\end{aligned}
\end{equation}
where $A$ represents the area of the entity region, which is  calculated by multiplying entity length  $b$ and the number of entity type $c$.  $C$ represents the center of entity region. Specially, the variable with an additional superscript \textit{$gt$} indicates that it is obtained from the annotations of the training dataset; otherwise, is obtained from the prediction of the model. The term $\rho^2(C, C^{gt})$  denotes the Euclidean distance between the center of predicted entity region $C$ and labeled entity region $C^{gt}$.  $V$ is an invariant to the regression scale and is normalized to the range $[0, 1]$, and $\frac{V^2}{(1-IoU) + V} $ serves as a special penalty when the predicted entity and ground truth entity are close but not match perfectly.

\paragraph{Entity Multi-Class Imbalance Loss.}

Inspired by multi-class loss \cite{Sun_2020_CVPR}, which is widely used in the text, image and numerical data,
we design the Entity Multi-class Cross-entropy (EMC) loss to  address the issue of entity imbalanced classification. 
During the loss computation, we iterate through each entity type in the pre-defined set $E$. For each iteration, the current entity type is designated as the positive class, while all other entity types are designated as the negative class. Based on the annotations in the training dataset, $\mathbb{P}_c$ includes all entities belonging to the positive class, and  $\mathbb{N}_c$  includes all entities belonging to the negative class.
To minimize the distance among entities in the positive samples and maximize the distance among entities in the negative samples, the EMC loss function ensures that: 

\begin{equation}
\renewcommand{\arraystretch}{0.8}
\begin{aligned}
    \mathcal{L}_{EMC} &= \sum_{c \in E} 
 \log{(1+\sum_{n \in \mathbb{N}_c} \mathrm{e}^{s_n} \sum_{m \in \mathbb{P}_c} \mathrm{e}^{-s_m})}, \\
\end{aligned}
\end{equation}
where  $s_m$ and $s_n$ are the span representation for positive entity and negative entity, respectively. 

Finally, we train the model by the EIoU loss and EMC loss. The total loss is computed as:

\begin{equation}
\renewcommand{\arraystretch}{0.8}
\mathcal{L} = \beta \mathcal{L}_{EIoU} + ( 1-\beta ) \mathcal{L}_{EMC}, 
\end{equation}
where $\beta$ is a hyperparameter of weight which is between EIoU loss and EMC loss.

\section{Experiments and Discussion}

\subsection{Dataset}\label{sec:data}

\begin{table}

\renewcommand{\arraystretch}{0.75}
\caption{Statistic of Datasets. The number following \textit{Entity} is the number of entity category in this dataset.}
\label{statistic}
\centering

\begin{tabularx}{\columnwidth}{@{\extracolsep{\fill}}llrrrr}
\toprule
\textbf{Dataset} & \textbf{Type} & \textbf{Train} & \textbf{Devel} & \textbf{Test} & \textbf{Total} \\
\midrule
\multirow{2}{*}{CMeEE} & \textit{Sentence} & 15000 & 2500 & 2500 &  20000 \\
& \textit{Entity} (9) &  82658 & 12671 & 13428 & 108757 \\
\midrule
\multirow{2}{*}{GENIA} & \textit{Sentence} & 15023 & 1669 & 1854 & 18546 \\
& \textit{Entity} (5) & 45041 & 4279 & 5351 & 54671 \\
\midrule
\multirow{2}{*}{ICEM} & \textit{Sentence} & 10452 & 1303 & 1359 & 13114 \\
& \textit{Entity} (71) & 41561 & 5163 & 5513 & 52237 \\
\bottomrule
\end{tabularx}
\end{table}

\begin{figure}
    \centering  
    \begin{minipage}{.5\columnwidth}  
        \centering
        \includegraphics[width=0.9\linewidth]{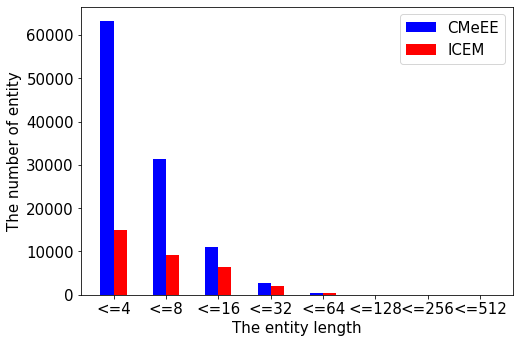}
        \subcaption{}
        \label{fig_entity_length}
    \end{minipage}%
    \begin{minipage}{.5\columnwidth}  
        \centering
        \includegraphics[width=0.9\linewidth]{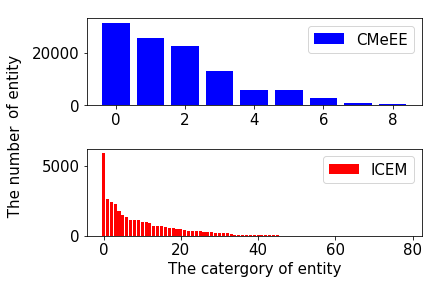}
        \subcaption{}
        \label{fig_entity_class}
    \end{minipage}  
    \caption{A comparison between CMeEE and ICEM datasets. 
    Fig. \ref{fig_entity_length} shows the distribution of entity lengths and the number of entities in two datasets. 
    Fig. \ref{fig_entity_class} illustrates the distribution of entity categories and the number of entities in each category in these datasets.
    }
    \label{fig:test}  
\end{figure}

We conduct experiments on three distinct datasets: CMeEE \cite{zhangCBLUEChineseBiomedical2022} and GENIA \cite{kimGENIACorpusSemantically2003} as the biomedical domain dataset and a new instrial complex equipment maintenance (ICEM) dataset as a low-resource dataset we collected from industrial application documents.
The ICEM dataset consists of 13,000 sentences, and the average length of entities in this dataset is completely different from the common NER datasets. The details of the datasets and the comparisons are shown in Table \ref{statistic} and Figure \ref{fig:test}.

\begin{table*}
\renewcommand{\arraystretch}{0.8}
 \caption{Analysis in low-resource scenarios on CMeEE dataset}
\label{analysis_of_cmeee}
\centering
\begin{tabularx}{0.9\textwidth}{@{\extracolsep{\fill}}lllllllll}
\toprule
\textbf{Entity}& \textbf{Method} & \textbf{1\%} & \textbf{2\%} & \textbf{5\%} & \textbf{10\%} & \textbf{20\%} & \textbf{50\%} & \textbf{100\%} \\
\midrule
\multirow{2}{*}{Boundary} & \citet{suGlobalPointerNovel2022} & 57.79 & 65.82 & 70.41 & 72.91 & 74.55 & 76.27 & 77.48 \\
& Ours & \textbf{60.19} & \textbf{67.48} & \textbf{71.05} & \textbf{73.85} & \textbf{75.16} & \textbf{76.82} & \textbf{77.68}  \\
\midrule
\multirow{2}{*}{Category} & \citet{suGlobalPointerNovel2022} & 52.25 & 60.53 & 64.29 & 68.33 & \textbf{70.48} & 72.27 & 73.79 \\
&Ours & \textbf{52.26} & \textbf{60.57} & \textbf{65.59} & \textbf{68.54} & 70.40 & \textbf{72.41} & \textbf{73.83} \\
\bottomrule
\end{tabularx}
\end{table*}

\textbf{CMeEE} dataset  \cite{zhangCBLUEChineseBiomedical2022} is a public dataset about Chinese medical nested NER dataset from medical textbooks, including 9 different entity categories. The dataset has a long-tail distribution which is challenging in NER task.

\textbf{GENIA} dataset \cite{kimGENIACorpusSemantically2003} is a biomedical nested NER dataset with characteristics of long-tail distribution of entity category, We preprocess the dataset following \citet{zheng2019boundary}.

\textbf{ICEM} dataset is an industrial domain NER dataset. The dataset is derived from industrial equipment operation and maintenance documents. It contains more than 70 different industrial domain entity categories. The dataset is a domain-specific low-resource dataset and there is a serious long-tail distribution among entities.

We made a comparison between Chinese CMeEE and ICEM datasets using statistical methods. As shown in Table \ref{statistic}, the datasets are close in terms of the number of sentences and entities. 
The sentences and entities in these datasets are similar in length distribution, it is shown in Fig \ref{fig_entity_length}. The average sentence lengths are 54 characters in CMeEE dataset and 56 characters in ICEM dataset. 
The average length of entity charater in CMeEE and ICEM datasets are 5 and 7, respectively. 
As the average length of entities increases, the recognition of entities becomes more difficult. 
Moreover, the number of entity categories have a significant difference between the two datasets. As shown in Fig \ref{fig_entity_class}, the CMeEE dataset has 9 different entities, their number decreases steadily. 
The ICEM datset has more than 80 classes, a few entity categories account for the majority of the number, most of the entity categories have little data. 
There are 55 kinds of entity quantity in ICEM dataset less than the smallest category in CMeEE dataset.

\begin{table}
\renewcommand{\arraystretch}{0.75}
\caption{Results of CMeEE, GENIA, and ICEM Datasets.}
\label{tab:result}
\centering
\begin{tabular}{lllrrr}
\toprule
\textbf{Dataset} & \textbf{Method} & \textbf{Precision} & \textbf{Recall} & \textbf{F1-Score}\\
\midrule
\multirow{6}{*}{\textbf{CMeEE}} &  \citet{guDelvingDeepRegularity2022}  & 66.25 & 64.89 & 65.57 \\ 
& \citet{yangTsERLTwostageEnhancement2022}  & 61.82 & 64.78 & 63.27 \\
& \citet{liuMedBERTPretrainingFramework2022a}  & 67.99 & 70.81 & 69.37 \\ 
& \citet{duMRCbasedMedicalNER2022} & 67.21 & 71.89 & 69.47 \\
& \citet{suGlobalPointerNovel2022}  & 73.35 & \textbf{74.24} & 73.79 \\
& Ours & \textbf{73.49} & 74.18 & \textbf{73.83} \\
\midrule
\multirow{4}{*}{\textbf{GENIA}} &
\citet{ebertsSpanBasedJointEntity2020} & 77.24 & 78.56 & 77.89 \\
& \citet{zhangSmartSpanNERMakingSpanNER2023} & 71.61 & \textbf{82.22} & 76.54 \\
& \citet{suGlobalPointerNovel2022} & \textbf{80.39} & 75.87 & 77.41 \\
& Ours & 79.70 & 77.02 & \textbf{78.34} \\
\midrule
\multirow{4}{*}{\textbf{ICEM}} &
\citet{zheng2019boundary} & 48.99 & 45.95 & 47.42 \\
& \citet{liUnifiedNamedEntity2021} & \textbf{65.16} & 46.74 & 54.44 \\ 
& \citet{suGlobalPointerNovel2022} & 57.83 & 54.38 & 56.04 \\
& Ours & 58.18 & \textbf{54.58} & \textbf{56.32} \\
\bottomrule
\end{tabular}
\end{table}

\begin{table}
\renewcommand{\arraystretch}{0.8}
\caption{Analysis of imbalanced classification on GENIA dataset}
\label{result_of_multitask}
\centering
\begin{tabularx}{\columnwidth}{@{\extracolsep{\fill}}lllcc}
\toprule
\multicolumn{2}{c}{\textbf{Category}}  & \multirow{2}{*}{\textbf{Ours}} & \multirow{2}{*}{\textbf{\citet{suGlobalPointerNovel2022}}} & \multirow{2}{*}{\textbf{Increase}} \\ 
Type & Ratio \\
\midrule
Protein & 60.7\% & 80.60 & 80.09 & +0.51 \\
DNA & 18.1\% & 76.20 & 74.53 &  +1.67 \\
Cell type & 12.7\% & 73.23 & 71.53 & +1.70 \\
Cell line & 6.8\% & 73.92 & 72.78 & +1.14 \\
RNA & 1.7\% & 86.91 & 84.97 & \textbf{+1.94} \\
\midrule
\multicolumn{2}{c}{Boundary} & 81.14 & 80.94 & +0.20 \\
\midrule
\multicolumn{2}{c}{All} & 78.34 & 77.41 & +0.93 \\
\bottomrule
\end{tabularx}
\end{table}

\subsection{Experimental Setup} \label{sec:setup}

In our experiment, we use precision (P), recall (R) and micro F1-score (F1) by micro average of these three runs by different random seeds.
We use the BERT-based model to encode raw text. Specifically, for the GENIA and NCBI datasets, we utilized BioBERT \cite{leeBioBERTPretrainedBiomedical2020} as the pre-trained language model encoder whereas for the CMeEE and ICEM Chinese datasets, we employed BERT-base-Chinese \cite{devlin-etal-2019-bert}.

\subsection{Main Results} \label{sec:main}

In this part, we show the results of the main experiments in Table \ref{tab:result}. Overall, our method has achieved performance on all three datasets. 

The results on CMeEE dataset is in Table \ref{tab:result}, \citet{guDelvingDeepRegularity2022},  \citet{yangTsERLTwostageEnhancement2022} and \citet{liuMedBERTPretrainingFramework2022a} span-based method reached the F1 score on 65.57, 63.27 and 69.73, respectively. Our proposed method achieves the best F1 score of 73.83, which outperformed 73.79 the F1 value of baseline model \cite{suGlobalPointerNovel2022} and the three span-based methods compared above.

We compare the span-based methods to our proposed method. \citet{ebertsSpanBasedJointEntity2020} proposed the F1 value of 77.89. \citet{zhangSmartSpanNERMakingSpanNER2023} elevate the F1 score on GENIA to 76.54. GP model \cite{suGlobalPointerNovel2022} achieves a 77.41 F1 score. We apply the EIoU-EMC loss to the GP  model, to outperform 0.93 on the F1 score.

Finally, the experiments are on the ICEM dataset. We evaluate the three different span-based models. As shown in Table \ref{tab:result}, \citet{zheng2019boundary} model proposed the F1 score of 47.42, \citet{liUnifiedNamedEntity2021}  achieves a 54.44 F1 score and baseline \cite{suGlobalPointerNovel2022} obtains the F1 score of 56.04. Our method reaches the best F1 score of 56.32 on this dataset.

\subsection{Analysis in Low Resource Scenario}\label{sec:low}
We further investigated the proposed model performance on a low-resource scenario. For this part of the experiment, we descend training data size on the CMeEE dataset, the largest dataset with most entities. We set the training set size to 1\%, 2\%, 5\%, 10\%, 20\%, and 50\% of the whole dataset, and conducted experiments on the baseline model and our proposed method. As shown in Table \ref{analysis_of_cmeee}, the results show our proposed method has achieved improvements in all sizes of boundary detection tasks, and the improvements are outstanding in the case of smaller-size data. For entity category classification, our proposed model proposed better or more competitive results than the baseline model in most cases. 

\subsection{Analysis in Imbalanced Classification} \label{sec:imb}
In this part, we chose the GENIA dataset for imbalanced classification experiments, because it contains an appropriate number of categories and a sufficient number of entities in each category.
To ensure the improvement of significant distinctions in performance among each task in our proposed method, we analyzed the evaluation metrics for each entity type and entity boundary as shown in Table \ref{result_of_multitask}. The middle part of this table displays the performance of the entity boundary recognition task. 
Among these five entity categories, we demonstrated through experiments that our proposed method is effective for the imbalanced classification subtask. As shown in Table \ref{result_of_multitask}, with the number of entities decreasing,
With the number of entities decreasing, the improvement of our proposed method in the F1 score becomes more significant. 
The entity category of RNA has a 1.94 increase in F1 score, of which accounts for only 1.7\% of all entity data, and it is the most significant improvement among all entity categories.

\section{Conclusion and Future Work}
This paper primarily introduces the proposed EIoU-EMC loss and validates the model's performance on biomedical datasets and the industrial application dataset we developed, demonstrating the effectiveness of this loss function in low-resource datasets and minority class samples in imbalanced classification tasks.
In future work, we will further explore the performance of the loss in other information retrieval tasks, while also refining the ICEM dataset to enhance its application value in knowledge-graph-based information retrieval.

\bibliographystyle{ACM-Reference-Format}
\bibliography{myref}

\end{document}